%% file: eccv2018submission.tex
\documentclass[runningheads]{llncs}

\usepackage{graphicx}
\usepackage[caption=false]{subfig}
\usepackage{amsmath,amssymb,color,bm} % define this before the line numbering.
\usepackage{booktabs}
\usepackage{xcolor}
\usepackage{url}
\usepackage[colorlinks=true,citecolor=green]{hyperref}
\usepackage[width=122mm,left=12mm,paperwidth=146mm,height=193mm,top=12mm,paperheight=217mm]{geometry}
\usepackage{tikz}

\newcommand{\eg}{{\it e.g.}}

\newcommand{\etc}{{\it etc.}}
\newcommand{\reals}{\mathbb{R}}

\begin{document}
% \linenumbers
\pagestyle{headings}
\mainmatter
\def\ECCV18SubNumber{}  % Insert your submission number here

\title{Weakly Supervised Representation Learning for Unsynchronized Audio-Visual Events} % Replace with your title

\titlerunning{Weakly Supervised Representation Learning for Unsynchronized AV Events}

\authorrunning{Parekh et al.}

\author{Sanjeel Parekh\textsuperscript{1,2}, Slim Essid\textsuperscript{1}, Alexey Ozerov\textsuperscript{2}, Ngoc Q. K. Duong\textsuperscript{2},\\Patrick P{\'e}rez\textsuperscript{2}, and Ga\"{e}l Richard\textsuperscript{1}}
\institute{\textsuperscript{1}LTCI, T{\'e}l{\'e}com ParisTech, Universit{\'e} Paris--Saclay, France \\ \textsuperscript{2}Technicolor, France}

\maketitle

\begin{abstract}
Audio-visual representation learning is an important task from the perspective of designing machines with the ability to understand complex events. To this end, we propose a novel multimodal framework that instantiates multiple instance learning. We show that the learnt representations are useful for classifying events and localizing their characteristic audio-visual elements. The system is trained using only video-level event labels without any timing information. An important feature of our method is its capacity to learn from unsynchronized audio-visual events. We achieve state-of-the-art results on a large-scale dataset of weakly-labeled audio event videos. Visualizations of localized  visual regions and audio segments substantiate our system's efficacy, especially when dealing with noisy situations where modality-specific cues appear asynchronously. 

\end{abstract}

%% Introduction

\section{Introduction}

We are surrounded by events that can be perceived via %composed of 
distinct audio and visual cues. Be it a ringing phone or a car passing by, we instantly identify the audio-visual (AV) components that characterize these events. This remarkable ability helps us understand and interact with our environment. For building machines with such scene understanding capabilities, it is important to design algorithms for learning audio-visual representations from real-world data. This work is a step in that direction, where we aim to learn such representations through weak supervision.

Specifically, we deal with the problem of event classification and characteristic audio-visual element localization in videos. Obtaining precisely annotated data for doing so is an expensive endeavor, made even more challenging by multimodal considerations. The annotation process is not only error prone and time consuming but also subjective to an extent. Often, event boundaries in audio, extent of video objects or even their presence is ambiguous. Thus, we opt for a weakly-supervised learning approach using data with only video-level event labels, that is labels given for whole video documents without timing information.

\begin{figure*}[t!]
	\centering
	\includegraphics[scale=0.3]{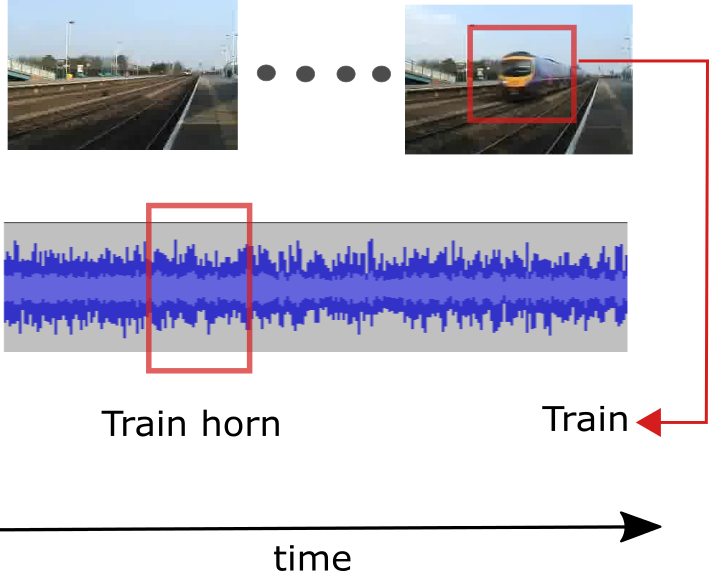}
	\caption{\textbf{Pictorial representation of the problem}: Given a video labeled as ``train horn'', we would like to: (i) identify the event, and (ii) localize both, its visual presence and the characteristic sound in the audio recording. Note that the train horn may sound before the train is visible. Our network can deal with such unsynchronized audio-visual events.}
\label{prob}
\end{figure*}

To motivate our tasks and method, consider a video labeled as ``train horn'', depicted in Fig. \ref{prob}. Assuming that the train is both visible and audible at some time in the video, in addition to identifying the event, we are interested in learning representations that help us answer the following questions:
\begin{itemize}
\item \textit{Where is the visual object or context that distinguishes the event?} In this case it might be the train (object) or tracks, platform (context) \etc{} We are thus aiming for their spatio-temporal localization in the image sequence.
\item \textit{When does the sound event occur?} Here it is the train horn. We thus want to temporally localize the audio event.
\end{itemize}

The variety of noisy situations that one may encounter in unconstrained environments or videos adds to the difficulty of this very challenging problem. Apart from modality-specific noise such as visual clutter, lighting variations and low audio signal-to-noise ratio, in real-world scenarios the audio and visual elements characterizing the event are often unsynchronized in time. This is to say that the train horn may sound before or after the train is visible, as in previous example. In the extreme, not so rare case, the train may not appear at all. We are interested in designing a system to tackle the aforementioned questions and situations.

Prior research has utilized audio and visual modalities for classification and localization tasks in various contexts. Fusing modality-specific hand-crafted or deep features has been a popular approach for problems such as multimedia event detection and video concept classification \cite{JiangCottonChangEtAl2009,ChangEllisJiangEtAl2007,jiang2018exploiting,jiang2013high}. On the other hand, audio-visual correlations have been utilized for localization and representation learning in general, through feature space transformation techniques such as canonical correlation analysis (CCA) \cite{IzadiniaSaleemiShah2013,KidronSchechnerElad2005} or deep networks \cite{owens2016visually,owens2016ambient,Arandjelovic17,Arandjelovic17b,andrew2013deep}. However, a unified multimodal framework for our task, that is learning data representations for simultaneously identifying real world events and the audio-visual cues depicting them has not been extensively studied in the literature.

Such tasks can be naturally interpreted as \textit{multiple instance learning} (MIL) problems \cite{dietterich1997solving}. MIL is typically applied to cases where labels are available over bags (sets of instances) instead of individual instances. The task then amounts to jointly selecting appropriate instances and estimating classifier parameters.  In our case, a video can be seen as a labeled bag, containing a collection of image regions (also referred to as \textit{image proposals}) and audio segments (also referred to as \textit{audio proposals}). This principle step is at the core of our approach. Interestingly, Jiang \textit{et al}. \cite{JiangCottonChangEtAl2009} deal with the broader task of video concept detection using an MIL formulation. Unlike us, they consider video segments to be bags of short-term audio-visual atoms (S-AVA). S-AVAs are multimodal feature vectors composed by concatenating hand-crafted appearance, motion and audio descriptors from image region trajectories and audio. Their method's reliance on a computationally expensive image segmentation procedure limits its application to large-scale datasets.

In this work, we decompose a video into image regions and temporal audio segments, dealing with each in separate visual and audio sub-modules. The key idea is to extract features from generated proposals and transform them for: (1) scoring each according to their relevance for class labels; (2) aggregating these scores in each modality and fusing them for video-level classification. This allows us to train both the sub-modules together through weak-supervision and learn representations for event classification and localization. Moreover, use of both the modalities makes the system robust against noisy scenarios.

Our main contributions are as follows: (1) we propose a new multimodal framework that allows jointly classifying videos and localizing both audio and visual cues responsible for this classification; (2) our approach, by construction, allows dealing with difficult cases when those cues are not synchronized in time; (3) we validate the system's performance, both quantitatively and qualitatively over a large-scale weakly-labeled dataset for audio events. State-of-the-art performance is achieved on the task of event classification. We also show, through a careful analysis of each sub-module, the useful complementary information held in each modality. Qualitative localization results confirm our technique's ability to identify event-specific AV cues. Moreover, localization in noisy situations, especially with regard to unsynchronized AV events, underlines our system's effectiveness.

We begin by discussing related work in the areas of computer vision, machine listening and multimodal representation learning in Section \ref{related}. This is followed by a detailed description of our pipeline for AV representation learning and classification, dealing with possibly asynchronous events in Section \ref{propapp}. Finally, we validate the usefulness of the learnt representations with a thorough analysis in Section \ref{expt}.

%% ------------Related Work-------------------
\section{Related work}
\label{related}
Researchers in computer vision and machine listening have independently applied several techniques for weakly supervised classification and localization of  visual objects and audio events, respectively. Recent progress in the area of multimodal deep representation learning has also led to several successful fusion-based approaches. To position our work, we briefly discuss most relevant developments in each of these domains.

\textbf{Object Localization and Classification.\enspace} There is a long history of works in computer vision applying weakly supervised learning for object
localization and classification. MIL techniques have been extensively used for this purpose \cite{bilen2014weakly,kantorov2016contextlocnet,bilen2016weakly,zhang2006multiple,cinbis2017weakly,oquab2015object,zhou2016learning}. Typically, each image is represented as a set of regions. Positive images contain at least one region from the reference class while negative images contain none. Latent structured output methods, \eg, based on support vector machines (SVMs)\cite{bilen2014object} or conditional random fields (CRFs)\cite{deselaers2010localizing}, address this problem by alternating between object appearance model estimation and region selection. Some works have focused on better initialization and regularization strategies \cite{song2014weakly,cinbis2017weakly,kumar2010self} for solving this non-convex optimization problem. 

Owing to the exceptional success of convolutional neural networks (CNNs) in computer vision, recently, several approaches have looked to build upon CNN architectures for embedding MIL strategies. These include the introduction of operations such as $\max$ pooling over regions \cite{oquab2015object}, global average pooling \cite{zhou2016learning} and their soft versions \cite{kolesnikov2016seed}. Another line of research consists in CNN-based localization over class-agnostic region proposals \cite{bilen2016weakly,kantorov2016contextlocnet,gkioxari2015contextual} extracted using a state-of-the-art proposal generation algorithm such as EdgeBoxes \cite{zitnick2014edge}, Selective Search \cite{uijlings2013selective} \etc{} These approaches are supported by the ability to extract fixed size feature maps from CNNs using region-of-interest \cite{girshick2015fast} or spatial pyramid pooling \cite{he2015spatial}. Our work is related to such techniques. We build upon ideas from two-stream architecture \cite{bilen2016weakly} for classification and localization.

\textbf{Audio Event Detection.\enspace} A significant amount of literature exists on supervised audio event detection (AED) \cite{mesaros2015sound,zhuang2010real,adavanne2017sound,bisot2017overlapping}. However, progress with weakly labeled data in the audio domain has been relatively recent. An early work \cite{kumar2016audio} showed the usefulness of MIL techniques to audio using SVM and neural networks.

The introduction of the weakly-labeled audio event detection task in the 2017 DCASE challenge\footnote{\url{http://www.cs.tut.fi/sgn/arg/dcase2017/challenge/}}, a challenge on Detection and Classification of Acoustic Scenes and Events, along with the release of Google AudioSet data\footnote{\url{https://research.google.com/audioset/}} \cite{gemmeke2017audio}, has led to accelerated progress in the recent past. AudioSet is a large-scale weakly-labeled dataset of audio events collected from YouTube videos. A subset of this data was used for the DCASE 2017 task on large-scale AED for smart cars \cite{DCASE2017challenge}.\footnote{\url{http://www.cs.tut.fi/sgn/arg/dcase2017/challenge/task-large-scale-sound-event-detection}} Several submissions to the task utilized sophisticated deep architectures with attention units \cite{Xu2017}, as well as $\max$ and softmax operations \cite{Salamon2017}. Another recent study introduced a CNN with global segment-level pooling for dealing with weak labels \cite{kumar2017knowledge}. While we share with these works the high-level goal of weakly-supervised learning, apart from our multimodal design, our audio sub-module, as discussed in the next section, is significantly different.

\textbf{Multimodal Deep Learning.\enspace} Lately, rapid progress in the application of deep learning methods to representation learning has motivated researchers to use them for fusing multimodal data.

The use of artificial neural networks for audio-visual fusion can be traced back to Yuhas \textit{et al}. \cite{yuhas1989integration}. The authors proposed to estimate acoustic spectral shapes from lip images for audio-visual speech recognition. In a later work, Becker and Hinton \cite{becker1992self} laid down the ideas for self organizing neural networks, where modules receiving separate but related inputs aim to produce similar outputs. Owing to the advent of large scale datasets and training capabilities, each of these formulations has recently emerged in broader contexts for audio-visual fusion in generic videos. Notably, Owens \textit{et al}. \cite{owens2016visually} train a CNN and Recurrent Neural Network (RNN) based architecture to predict audio using visual input. In another work, this idea is extended to predict the audio category of static images \cite{owens2016ambient}. Transfer learning experiments on image classification confirm that ambient audio assists visual learning \cite{owens2016ambient}. This is  reversed by \cite{aytar2016soundnet} to demonstrate how audio representation learning could be guided through visual object and scene understanding using a teacher-student learning framework. Herein the authors use trained visual networks to minimize the Kullback-Leibler divergence between the outputs of visual and audio networks. The resulting features are shown to be useful for audio event detection tasks. Subsequently, these ideas were extended to learning shared audio-visual-text representations \cite{aytar2017see}.

In some very recent works, useful audio-visual representations are learnt through the auxiliary task of training a network to predict audio-visual correspondence \cite{Arandjelovic17,Arandjelovic17b}. Indeed, the learnt audio representations in \cite{Arandjelovic17} achieve state-of-the-art results on audio event detection experiments. By design, such systems do not deal with the case of unsynchronized AV events as discussed earlier. Other notable approaches include multimodal autoencoder architectures \cite{ngiam2011multimodal} for learning shared representations even for the case where only a single view of data is present at training and testing time. Another interesting work extends CCA to learning two deep encodings, one for each view, such that their correlation is maximized \cite{andrew2013deep}.

Our work is significantly different from earlier studies on several counts: Contrary to prior work, where unsupervised representations are learnt through audio--image correlations, we adopt a weakly-supervised learning approach using event classes. Unlike \cite{Arandjelovic17b,owens2016visually,owens2016ambient}, we focus on localizing discriminative audio and visual components for real-world events. We formulate the problem as a multiple instance learning task using class-agnostic proposals from both video frames and audio.  This allows us to simultaneously solve the classification and localization problems. Most importantly, by construction, our framework deals with the difficult case of asynchronous audio-visual events.

%% --------------- Our approach------------

\begin{figure*}[t!]
	\label{prosys}
	\centering
 \input{block_diagram.tex}
	\caption{\textbf{Proposed approach}: Given a video, we consider the depicted pipeline for going from audio and visual proposals to localization and classification. Here $\bm W_{\text{cls}}$ and $\bm W_{\text{loc}}$ refer to the fully-connected classification and localization streams respectively; $\sigma$ denotes softmax operation over proposals for each class, $\odot$ refers to element-wise multiplication; $\Sigma$ to a summation over proposals and $\ell_2$ to a normalization of scores. }
\end{figure*}
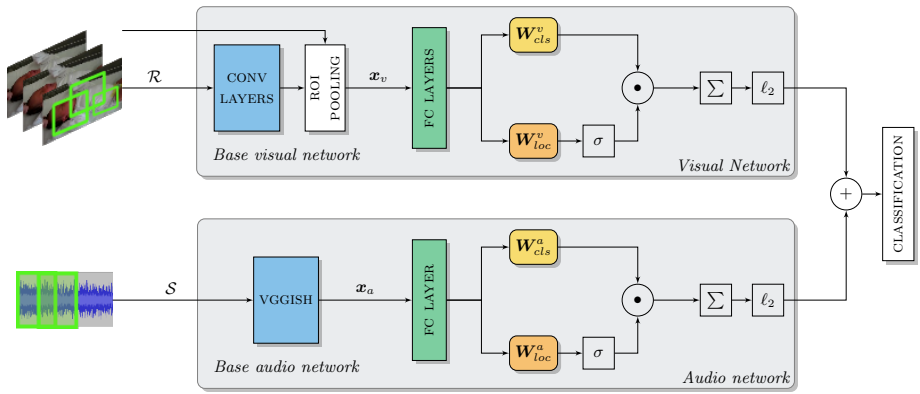

\section{Proposed Approach}
\label{propapp}
 An overview of our approach is provided in Fig. \ref{prosys}. We model a video $V$ as a bag of $M$ selected image regions, $\mathcal{R} = \{r_1, r_2, \ldots,r_M\}$, obtained from sub-sampled frames and $T$ audio segments, $\mathcal{S} = \{s_1, s_2, \ldots, s_T\}$. Given $N$ such training examples, $\mathcal{V} = \{V^{(n)}\}_{n=1}^{N}$, organized into $C$ classes, our goal is to learn a representation to jointly classify and localize image regions and audio segments that characterize a class. We begin by computing features over proposals in respective modalities, which are then passed through independent scoring networks. Finally, audio and visual sub-module scores are combined for classification. Each of these components is discussed below in detail.

\subsection{Generating and Extracting Features from Proposals}

\textbf{Visual Proposals.\enspace} Generating proposals for object containing regions from images is at the heart of various visual object detection algorithms \cite{girshick2014rich,wang2013regionlets}. As our goal is to spatially and temporally localize the most discriminative region pertaining to a class, we choose to apply this technique over sub-sampled video frame sequences. In particular, we sub-sample the extracted frames of each video at a rate of 1 frame per second. This is followed by class-agnostic region proposal generation on the sub-sampled images using EdgeBoxes \cite{zitnick2014edge}. This proposal generation method builds upon the insight that the number of contours entirely inside a box is indicative of the likelihood of an object's presence. Its use in our pipeline is motivated by experiments confirming better performance in terms of speed/accuracy tradeoffs over most competing techniques \cite{hosang2014good}. EdgeBoxes additionally generates a confidence score for each bounding box which reflects the box's ``objectness". To reduce the computational load and redundancy, we use this score to select the top $M_{\rm img}$ proposals from each sampled image, $I$, and use them for feature extraction. Hence, given a 10 second video, the aforementioned procedure would leave us with a list of $M = 10 \times M_{\rm img}$ region proposals.

A fixed-length feature vector, $\bm x_v(r_m;V)  \in \reals^{d_v}$ is obtained from each image region proposal, $r_m$ in $V$,  using a convolutional neural network altered with a region-of-interest (RoI) pooling layer. RoI
layer works by computing fixed size feature maps (\eg{} $6 \times 6$ for \texttt{caffenet} \cite{krizhevsky2012imagenet}) from regions of an image using max-pooling \cite{girshick2015fast}. This helps ensure compatibility between convolutional and fully connected layers of a network when using regions of varying sizes. Moreover, unlike Region-based CNN (RCNN) \cite{girshick2014rich}, shared computation for different regions of the same image using Fast-RCNN implementation \cite{girshick2015fast} leads to faster processing. In Fig. \ref{prosys} we refer to this feature extractor as the base visual network. In practice, feature vectors $\bm x_v(\cdot)$ are extracted after RoI 
pooling layer and passed through two fully connected layers, which are fine-tuned during training. Typically, standard CNN architectures pre-trained on ImageNet \cite{deng2009imagenet} classification are used for the purpose of initializing network weights.

\textbf{Audio Proposals.\enspace} We first represent the raw audio waveform  as a log-Mel spectrogram. Each proposal is then obtained by sliding a fixed-length window over the obtained spectrogram along the temporal axis. The dimensions of this window are chosen to be compatible with the audio feature extractor. For our system we set the proposal window length to 960ms and stride to 480ms. 

We use a  VGG-style deep network known as \texttt{vggish} for base audio feature extraction. Inspired by the success of CNNs in visual object recognition Hershey \textit{et al}. \cite{hershey2017cnn} introduced this state-of-the-art audio feature extractor as an audio parallel to networks pre-trained on ImageNet for classification. \texttt{vggish} has been pre-trained on a preliminary version of YouTube-8M \cite{45619} for audio classification based on video tags. It stacks 4 convolutional and 2 fully connected layers to generate a 128 dimensional embedding, $\bm x_a (s_t;V) \in \reals^{128}$ for each input log-Mel spectrogram segment $s_t \in \reals^{96 \times 64}$ with 64 Mel-bands and 96 temporal frames. Prior to proposal scoring, the generated embedding is passed through a fully-connected layer that is learnt from scratch.

\subsection{Proposal Scoring Network and Fusion}
So far, we have extracted base features for each proposal in both the modalities and passed them through fully connected layers in their respective modules.
Equipped with this transformed representation of each proposal, we use the two-stream architecture proposed by Bilen \textit{et al}. \cite{bilen2016weakly} for scoring each of them with respect to the classes. There is one scoring network of the same architecture for each modality as depicted in Fig. \ref{prosys}. Thus, for notational convenience, we generically denote the set of audio or visual proposals for each video by $\mathcal{P}$ and let proposal representations before the scoring network be stacked in a matrix $Z \in \reals^{|\mathcal{P}| \times d}$, where $d$ denotes the dimensionality of the audio/visual proposal representation.

The architecture of this module consists of parallel classification and localization streams. The former classifies each region by passing $Z$ through a linear fully connected layer with weights $W_{\text{cls}}$, giving a matrix $A \in \reals^{|\mathcal P| \times C}$. On the other hand, the localization layer passes the same input through another fully-connected layer with weights $W_{\text{loc}}$. This is  followed by a softmax operation over the resulting matrix $B \in \reals^{|\mathcal P| \times C}$ in the localization stream. The softmax operation on each element of $B$ can be written as:

\begin{equation}
[\sigma(B)]_{pc} = \frac{e^{b_{pc}}}{\sum_{p=1}^{|\mathcal{P}|} e^{b_{pc}}},~\forall (p,c) \in (1,|\mathcal{P}|)\times(1,C).
\end{equation}

This allows the localization layer to choose the most relevant proposals for each class. Subsequently, the classification stream output is weighted by $\sigma(B)$ through element-wise multiplication: $D = A \odot \sigma(B)$ . Class scores over the video are obtained by summing the resulting weighted scores in $D$ over proposals. Note that the dataset we use, by construction, allows a region or segment to belong to multiple classes. Hence, we do not opt for softmax on the classification stream, as done in \cite{bilen2016weakly}.

After performing the above stated operations for both audio and visual sub-modules, in the final step, the global video-level scores are $\ell_2$ normalized and added. In preliminary experiments we found this to work better than addition of unnormalized scores. We hypothesize that the system trains better because $\ell_2$ normalization ensures that the scores being added are in the same range.

\subsection{Classification Loss and Network Training} Given a set of $N$ training videos and labels, $\{V^{(n)}, \bm y^{(n)}\}$, we solve a multi-label classification problem. Here $\bm y \in \mathcal{Y} = \{-1,+1\}^C$ with the class presence denoted by +1 and absence by $-1$. 
To recall, for each video $V^{(n)}$, the network takes as input a set of image regions $\mathcal{R}^{(n)}$ and audio segments $\mathcal{S}^{(n)}$. After performing the described operations on each modality separately, the $\ell_2$ normalized scores are added and represented by $\phi(V^{(n)};\bm w) \in \reals^C$, with all network weights and biases denoted by $\bm w$. All the weights including and following fully-connected layer processing stage for both the modalities are included in $\bm w$. Note that both sub-modules are trained jointly. The network is trained using the multi-label hinge loss:

\begin{equation}
L(\bm w) = \frac{1}{CN}\sum_{n=1}^N \sum_{c=1}^C \max\Big(0, 1-y^{(n)}_c\phi_c(V^{(n)};\bm w)\Big).
\end{equation}

%%-------Experiments---------------

\section{Experimental Validation}
\label{expt}
\textbf{Dataset.\enspace}
We use the recently introduced dataset for DCASE challenge on large-scale weakly supervised sound event detection for smart cars \cite{DCASE2017challenge}. This is a subset of Audioset \cite{gemmeke2017audio} which contains a collection of weakly-annotated unconstrained YouTube videos of vehicle and warning sounds spread over 17 classes. It is categorized as follows:

\begin{itemize}
\item \textit{Warning sounds}: Train horn, Air horn, Truck horn, Car alarm,
Reversing beeps, Ambulance (siren), Police car (siren), Fire
engine fire truck (siren), Civil defense siren, Screaming.
\item \textit{Vehicle sounds}: Bicycle, Skateboard, Car, Car passing by, Bus,
Truck, Motorcycle, Train.
\end{itemize}

This multi-label dataset contains 51,172 training samples, 488 validation and 1103 testing samples. Despite our best efforts, due to YouTube and video downloader issues, some videos were unavailable, not downloaded or contained no audio. This left us with 48,715 training, 462 validation and 1103 testing clips.  It is worth mentioning that the training data is highly unbalanced with the number of samples for the classes ranging from 175 to 24K. To mitigate the negative effect of this imbalance on training, we introduce some balance by ensuring that each training batch contains at least one sample from some or all of the under-represented classes.  Briefly, each batch is generated by first randomly sampling labels from a specific list, followed by fetching examples corresponding to the number of times each label is sampled. This list is generated by ensuring higher but limited presence of classes with more examples. We use a publicly available  implementation for this purpose \cite{Xu2017}. \footnote{\url{https://github.com/yongxuUSTC/dcase2017_task4_cvssp/blob/master/data_generator.py}}

\textbf{Baselines.\enspace} To our knowledge, there is no prior work on deep architectures that perform the task of weakly supervised classification and localization for unsynchronized audio-visual  events. Our task and method are substantially different from recently proposed networks like L3 \cite{Arandjelovic17,Arandjelovic17b} which are trained using synchronous AV pairs on a large collection of videos in a self-supervised manner. However, we designed several strong baselines for comparison and an ablation study. In particular, we compare against the following networks:

\begin{enumerate}
\setlength\itemsep{1em}
\item AV One-Stream Architecture: Applying MIL in a straight-forward manner, we could proceed only with a single stream. That is, we can use the classification stream followed by a max operation for selecting the highest scoring regions and segments for obtaining global video-level scores. As done in \cite{bilen2016weakly}, we choose to implement this as a multimodal MIL-based baseline. We replace the $max$ operation by the \textit{log-sum-exponential} operator, its soft approximation. This has been shown to yield better results \cite{bilen2014weakly}. The scores on both the streams are $\ell_2$ normalized before addition for classification. This essentially amounts to removing from Fig. \ref{prosys} the localization branches and replacing the summation over proposals with the soft-maximum operation described above. To avoid any confusion, please note that we use the term `stream' to refer to classification and localization parts of the scoring network.

\item Visual-Only and Audio-Only Networks: These networks only utilize one of the modalities for classification. However, note that there are still two streams for classification and localization, respectively. For a fair comparison and ablation study we train these networks with $\ell_2$ normalization. In addition, for completeness we also implement Bilen \textit{et al}.'s architecture for weakly supervised deep detection networks (WSDDN)  with an additional softmax on the classification stream. As the scores are in the range [0,1], we train this particular network with $C$ binary log-loss terms \cite{bilen2016weakly}. When discussing results we refer to this system as WSDDN-Type.

\item CVSSP Audio-Only \cite{Xu2017}: This state-of-the-art method is the DCASE 2017 challenge winner for the audio event classification sub-task. The system is based on Gated convolutional RNN (CRNN) for better temporal modeling and attention-based localization. They use no external data and training/evaluation is carried out on all the samples. We present results for both their winning fusion system, which combines prediction of various models and Gated-RCNN model trained with log-Mel spectrum.
\end{enumerate}

\textbf{Implementation Details.\enspace} All systems, including variants, are implemented in Tensorflow. They were trained for 25K iterations using Adam optimizer \cite{adam} with a learning rate of $10^{-5}$ and a batch size of 24. We use the \textsc{matlab} implementation of EdgeBoxes for generating region proposals, obtaining approximately 100 regions per video with $K=10$ and a duration of 10 sec. The implementation is used with default parameter setting. Base visual features, $\bm x_v \in \reals^{9216}$ are extracted using \texttt{caffenet} with pre-trained ImageNet weights and RoI pooling layer modification. With $6\times 6$ RoI pooling we get a 9216 ($=256 \times 6 \times 6$)  dimensional feature vector. For this, the Fast-RCNN Caffe implementation is used \cite{girshick2015fast}.  The fully connected layers, namely $fc_6$ and $fc_7$, each with 4096 neurons, are fine-tuned, with 50\% dropout during training.

For audio, each recording is resampled to 16 kHz before processing. Log-Mel spectrum over the whole file is computed with a window size of 25ms and 10ms hop length. The resulting spectrum is chunked into segment proposals using a 960ms window with a 480ms stride. Note that the window and hop-length used for log Mel-spectrum computation is different from the one used for segment proposal extraction. For a 10 second recording, this yields 20 segments of size $96 \times 64$. We use the official Tensorflow implementation of \texttt{vggish}.\footnote{\url{https://github.com/tensorflow/models/tree/master/research/audioset}} The base audio features extracted from \texttt{vggish} are run through a fully connected layer with 128 neurons. This layer is learnt from scratch along with the scoring networks during training.

\textbf{Metrics.\enspace} The baselines and proposed systems are evaluated on the micro-averaged F1 score. The term micro-averaging implies that the F1 score is computed using a global count of total true positives, false negatives and false positives. This was the official metric used by DCASE 2017 smart cars task for ranking systems. The score thresholds for each system are determined by tuning over validation data to maximize F1 score for each class. They are then applied to the test data for final predictions. For further insight, we also report here the F1 scores for each class.

\subsection*{Results and Discussion}

\textbf{Quantitative Results. \enspace}We show in Table \ref{f1micro} 
the micro-averaged F1 scores for each of the systems described in the paper. In particular, systems (a)-(e) in Table \ref{f1micro} present various baselines (and also variants) of our audio-visual two-stream approach, (f)-(g) 
denote results from CVSSP team \cite{Xu2017}, winners of the DCASE AED for smart cars audio event tagging task. 
We outperform all the approaches by a significant margin. Among the multimodal systems, the two-stream architecture performs much better than the one-stream counter-part, designed with only a classification stream and soft-maximum for region selection. On the other hand, the state-of-the-art CVSSP fusion system, which combines predictions of various models, achieves a better precision than the other methods. Several important and interesting observations can be made by looking at these results in conjunction with the class-wise scores reported in Table \ref{clswise}.

Most importantly, the results emphasize the complementary role of visual and audio sub-modules for this task. To see this, we could categorize the data into two sets: (i) classes with clearly defined audio-visual elements, for instance car, train, motorcycle; (ii) some warning sounds such as, \eg{}, reverse beeping, screaming, air horn, where the visual object's presence is ambiguous. The class-wise results of the video only system are a clear indication of this split. 
Well-defined visual cues enhance the performance of the proposed multimodal system over audio-only approaches, as video frames carry vital information about the object. On the other hand, in the case of warning sounds, frames alone are insufficient as evidenced by results for the video-only system. In this case, the presence of audio assists the system in arriving at the correct prediction. The expected audio-visual complementarity is clearly established through these results. 

Note that for some warning sounds the CVSSP method achieves better results. In this regard, we believe better temporal modeling for our audio system could lead to further improvements. Particularly, currently we operate with a coarse temporal window of 960ms, which might not be ideal for all audio events. RNNs could also be used for further improvements. We believe such improvements are orthogonal and were not the focus of this study. We also observe that results for under-represented classes in the training data are relatively lower. This can possibly be mitigated through data augmentation strategies.

\begin{table}[!t]
\setlength{\tabcolsep}{12pt}
	\centering
	\begin{tabular}{r l c c c} 
		\toprule
		&System & F1 & Precision & Recall \\ [0.5ex] 
		\midrule
		(a)&Proposed AV Two Stream& \textbf{64.2} & 59.7 & \textbf{69.4}  \\ 
		(b)&TS Audio-Only & 57.3  & 53.2 &  62.0 \\
		
	(c)&TS Video-Only & 47.3 & 48.5 & 46.1\\
    (d)&TS Video-Only WSDDN-Type \cite{bilen2016weakly} & 48.8 & 47.6 & 50.1\\
		\midrule	
		(e)&AV One Stream & 55.3  & 50.4 & 61.2 \\
		\midrule

		(f)&CVSSP - Fusion system \cite{Xu2017} & 55.6 & \textbf{61.4} & 50.8 \\
        (g)&CVSSP - Gated-CRNN-logMel \cite{Xu2017}&  54.2 & 58.9 & 50.2 \\
		\bottomrule\\
	\end{tabular}
	\caption{Results on DCASE smart cars task test set. We report here the micro-averaged F1 score, precision and recall values and compare with state-of-the-art. TS is an acronym for two-stream.}
    \label{f1micro}
\end{table}

\begin{table}[!t]
\setlength{\tabcolsep}{5pt}
\centering
\resizebox{\textwidth}{!}{%
	\begin{tabular}{l@{\hspace{30pt}} c c c c c c c c c c c c c c c c c} 
\toprule
\multicolumn{1}{l}{System} & \multicolumn{8}{c}{Vehicle Sounds}&\multicolumn{9}{c}{Warning Sounds}\\
\cmidrule(lr){2-9}
\cmidrule(lr){10-18}
% & bik& bus&car & car-pby& mbik& skt& trn& trk& air-hrn& amb& car-alm& civ-def& f-eng& pol-car& rv-bps& scrm& trn-hrn\\ [0.5ex] 
%\midrule
%Proposed AV TS& \textbf{75.7} & \textbf{54.9}& \textbf{75.0} & \textbf{34.5} & \textbf{76.2} &\textbf{78.6} &82.0 & \textbf{61.5} &40.0 &\textbf{64.7} &53.9 &80.4 &64.4 &49.2 &36.6 &81.1 & 47.1  \\ [0.5ex]
%		TS Audio-Only & 42.1& 38.8& 69.8& 29.2& 68.9& 64.9& 78.5& 44.0& 40.0&58.2 &53.0 &79.6 &61.0 &51.4 &42.9 &72.1 & 46.9 \\ [0.5ex]
%		
%		TS Video-Only & 72.5& 52.0& 61.2&14.9 &54.1 &64.2 & 73.3& 49.7&12.0 & 33.9& 13.5& 68.6& 46.5&19.8 &21.8 &44.1 & 32.1 \\ [0.5ex]
%		\midrule 
%		AV OS  & 68.2& 53.6& 74.1& 25.6& 67.1& 74.4& \textbf{82.8}& 52.8& 28.0& 54.6& 20.6& 76.6& 60.4& 56.3& 18.8& 49.4& 36.2 \\ [0.5ex]
%		\midrule
%
%		CVSSP - FS &40.5 &39.7 &72.9 &27.1 &63.5 &74.5 &79.2 &52.3 & \textbf{63.7} &35.6 &\textbf{72.9} &\textbf{86.4} & \textbf{65.7} &\textbf{63.8} & \textbf{60.3} & \textbf{91.2}&\textbf{73.6}   \\ [0.5ex]
%		\bottomrule\\
& bik& bus&car & car-pby& mbik& skt& trn& trk& air-hrn& amb& car-alm& civ-def& f-eng& pol-car& rv-bps& scrm& trn-hrn\\ [0.5ex] 
\midrule
Proposed AV TS& \textbf{75.7} & \textbf{54.9}& \textbf{75.0} & \textbf{34.6} & \textbf{76.2} &\textbf{78.6} & 82.0 & \textbf{61.5} &40.0 &\textbf{64.7} &53.9 &80.4 &64.4 &49.2 &36.6 &81.1 & 47.1  \\ [0.5ex]
TS Audio-Only & 42.1& 38.8& 69.8& 29.6& 68.9& 64.9& 78.5& 44.0& 40.4&58.2 &53.0 &79.6 &61.0 &51.4 &42.9 &72.1 & 46.9 \\ [0.5ex]

TS Video-Only & 72.5& 52.0& 61.2&15.0 &54.1 &64.2 & 73.3& 49.7&12.0 & 33.9& 13.5& 68.6& 46.5&19.8 &21.8 &44.1 & 32.1 \\ [0.5ex]
\midrule 
AV OS  & 68.2& 53.6& 74.1& 25.6& 67.1& 74.4& \textbf{82.8} & 52.8& 28.0& 54.7& 20.6& 76.6& 60.4& 56.3& 18.8& 49.4& 36.2 \\ [0.5ex]
\midrule

CVSSP - FS &40.5 &39.7 &72.9 &27.1 &63.5 &74.5 &79.2 &52.3 & \textbf{63.7} &35.6 &\textbf{72.9} &\textbf{86.4} & \textbf{65.7} &\textbf{63.8} & \textbf{60.3} & \textbf{91.2}&\textbf{73.6}   \\ [0.5ex]
\bottomrule\\
	\end{tabular}}
    
	\caption{Class-wise comparison on test set using F1 scores. We use TS, OS and FS as acronyms to refer to two-stream, one-stream and fusion system, respectively.  }
    \label{clswise}
\end{table}

\begin{figure}[htp]
\centering
\subfloat[Train]{%
  \includegraphics[clip,width=\columnwidth]{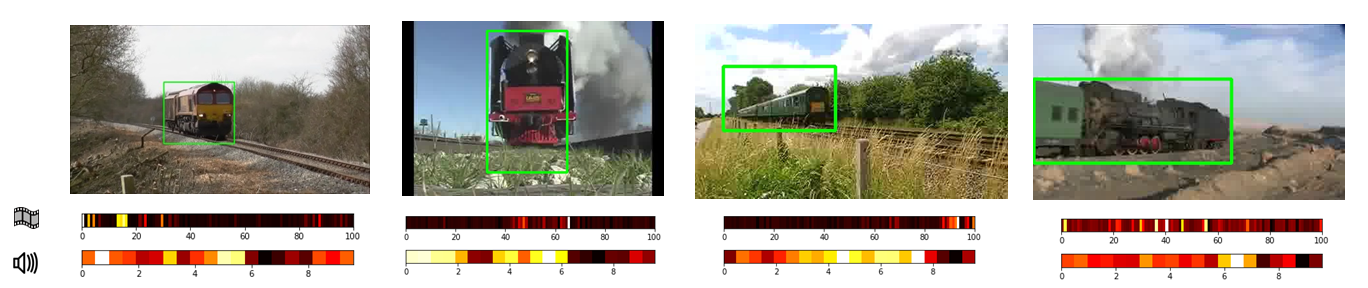}%
}

\subfloat[Bicycle]{%
  \includegraphics[clip,width=\columnwidth]{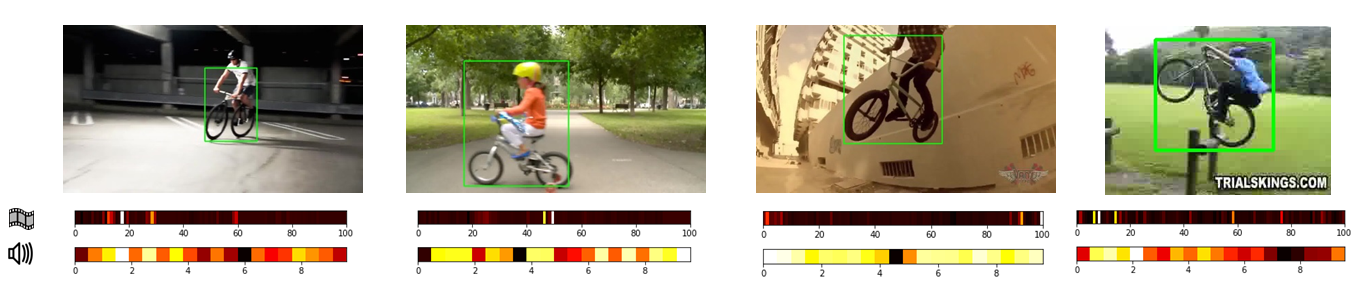}%
}

\subfloat[Car]{%
  \includegraphics[clip,width=\columnwidth]{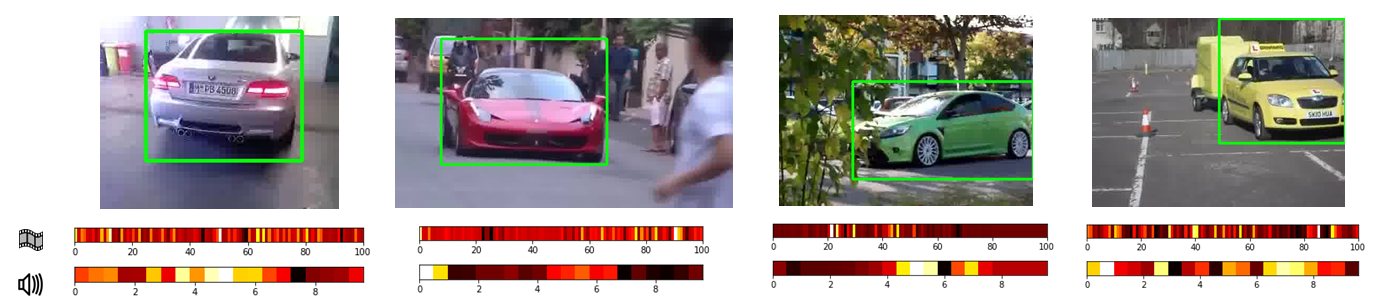}% 
}

\subfloat[Truck]{%
 \includegraphics[clip,width=\columnwidth]{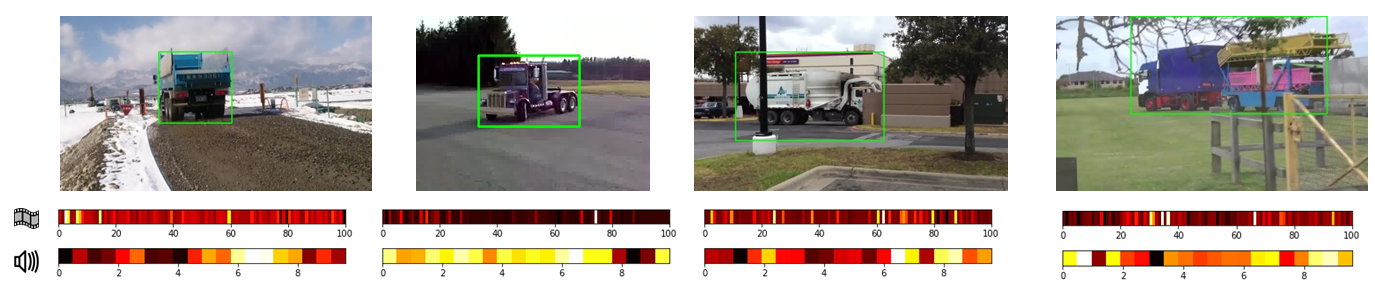}%
 } 

\subfloat[Motorcycle]{%
\includegraphics[clip,width=\columnwidth]{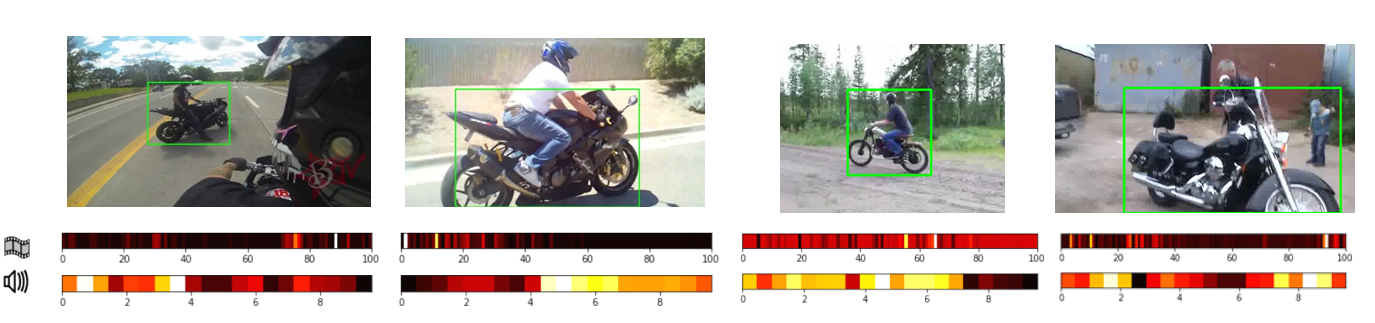}% 
}

\caption{Examples of localization on video frames for a few categories from the test data. The localization results are shown in green. Below each image we display the scaled region proposal (top) and audio segment scores for labels referred to in the caption. The visual heatmap is a concatenation of proposals from all the sub-sampled frames, arranged in temporal order.}
\label{ploc}
\end{figure}

\begin{figure}[!t]
\centering
\includegraphics[width=\textwidth]{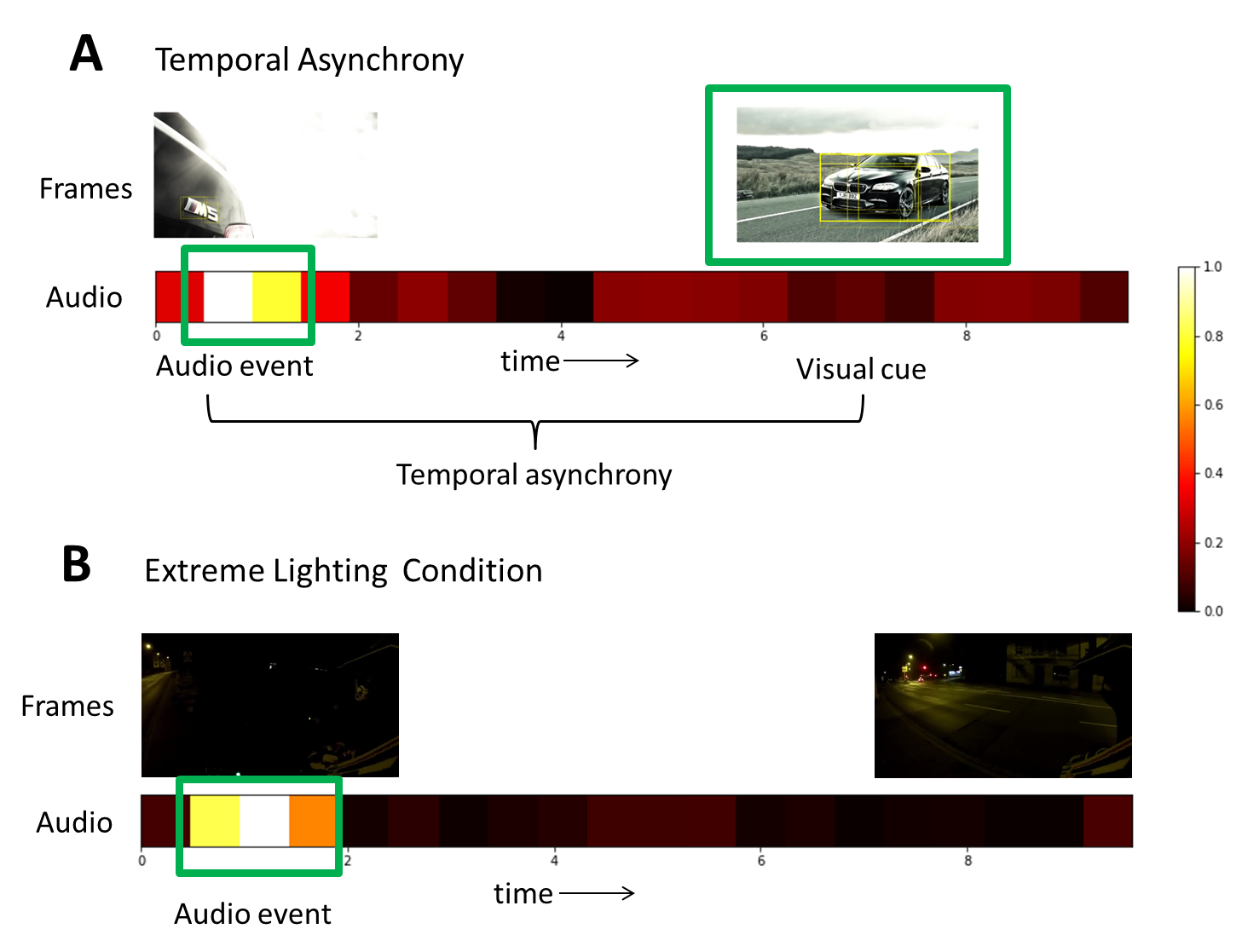}
\caption{Qualitative results for unsynchronized AV events. For both the cases A and B, the heatmap at the bottom denotes audio localization over segments for the class under consideration. For heatmap display, the audio localization vector has been scaled to lie between [0,1]. The top row depicts video frames roughly aligned to the audio temporal axis. (A) Top: Here we show a video where the visual object of interest appears after the audio event. This is a `car' video from the validation split. The video frames show bounding boxes where edge opacity is controlled by the box's detection score. In other words, higher score implies better visibility (B) Bottom:  This is a case from the evaluation data where due to lighting conditions, the visual object is not visible. However the system correctly localizes in audio and predicts the `motorcycle' class.}
\label{asyncloc}
\end{figure}

\textbf{Qualitative Results. \enspace} Fig. \ref{ploc} displays several video frames from different evaluation videos where we achieve good visual localization for various objects. The heatmaps shown below the images denote image region (top) and audio segment detection scores (bottom) for the reference class in sub-figure captions. The $x$-axis for the former denotes all the proposals from the subsampled images arranged in temporal order, whereas for audio it denotes the overlapping segment time-stamps. The display uses `hot' colormap where black is 0 and white 1, as depicted in Fig. \ref{asyncloc}. We see that the discriminative proposals are found at different time instants for each modality.

A by-product of our design is the ability to deal with asynchronous audio-visual events. We present in Fig. \ref{asyncloc} two examples to demonstrate this. In the first case \textbf{A}, the sound of a car's engine is heard in the first two seconds followed by music. The normalized audio localization heatmap at the bottom displays the scores assigned to each temporal audio segment, $s_t$ by the car classifier. The video frames placed above are roughly aligned with the audio temporal axis to show the video frame at the instant when the car sounds and the point where the visual network localizes. The localization is displayed through a yellow bounding box. To better understand the system's output, we modulate the opacity of the bounding box according to the system's score for it. Higher the score, more visible the bounding box. 
As expected, we do not observe any yellow edges in the first frame. Clearly, there exists temporal asynchrony, where the system locks onto the car, much later, when it is completely visible. \textbf{B} depicts an example, where due to extreme lighting conditions the visual object is not visible. Here too, we localize the audio object and correctly predict the `motorcycle' class.\footnote{Localization examples with audio can be found at \url{https://youtu.be/C-jrZ9SDMDY}}

\section{Conclusion}

We have proposed a novel approach based on a deep multimodal architecture for audio-visual events localization and classification. A particular strength of our system is its capability to deal with asynchronous audio-visual events for which typical visual and audio cues appear at different time instants. The proposed experiments have demonstrated the merits of our approach compared to several benchmark methods but have also shown that a more accurate audio temporal modeling would be needed to better cope with situations where the visual modality is inefficient.

\bibliographystyle{splncs}

\end{document}

%% file: block_diagram.tex
\usetikzlibrary{positioning,arrows,fit,calc,shadows,shapes,backgrounds}

\definecolor{myblue}{HTML}{85C1E9}
\definecolor{myyellow}{HTML}{F7DC6F}
\definecolor{mygreen}{HTML}{7DCEA0}%{7ABA43}%{E7F2C9}
\definecolor{fitcolor}{HTML}{EAECEE}
\definecolor{myred}{HTML}{F7C171}

\tikzset{%
	loc/.style={fill=myred},
	traintest/.style={fill=myblue},
	fixed/.style={fill=myblue},
	test/.style={fill=myyellow},
	train/.style={fill=mygreen},
	ops/.style={fill=white},
	fit node/.style={fill=fitcolor, drop shadow},
	block/.style={draw, fill=blue!20!white, rectangle,
		minimum height=7em, minimum width=2em, font=\footnotesize\scshape, align=center, drop shadow},
	blocknet/.style={draw, fill=blue!20!white, rectangle, minimum height=5em, minimum width=3em, font=\footnotesize\scshape, align=center, drop shadow},
	blocknetop/.style={draw, fill=blue!20!white, rectangle, minimum height=2em, minimum width=5em, font=\footnotesize\scshape, align=center, drop shadow},
	streamblock/.style={draw, fill=blue!20!white, rounded corners, rectangle,
		minimum height=2em, minimum width=2em, font=\footnotesize\scshape, align=center, drop shadow},
	W/.style={draw, rectangle, minimum height=7em, minimum width=1.5em, drop shadow, traintest},
	bbox/.style={draw=gray, rounded corners, rectangle},
	activ/.style={font=\tiny\itshape,at start,inner sep=2pt},
	fit label/.style={inner sep=.3em,font=\small\itshape},
	input/.style={inner sep=0pt},
	output/.style={inner sep=0pt},
	sum/.style = {draw, fill=white, circle, minimum size=.8em, node distance=5em, inner sep=0pt,traintest},
	pinstyle/.style = {pin edge={to-,thin,black}}
}

\tikzstyle{place}=[circle,draw,inner sep=0pt,minimum size=6mm]
\tikzstyle{transition}=[rectangle,draw,inner sep=0pt,minimum size=6mm]
\tikzstyle{transitioncls}=[rectangle,draw,inner sep=4pt,minimum size=6mm, drop shadow, fill=white]

\resizebox*{\textwidth}{!}{
\begin{tikzpicture}[>=latex', font=\small]
%Visual Net
\node at (-9,1) [inner sep=0pt] (vis) 
{\includegraphics[width=.18\textwidth]{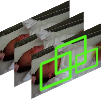}};
\node at (-5.5,1) [blocknet,fixed] (conv) {conv \\ layers};
\node at (-4,1) [blocknetop,ops,rotate=90] (roi) {roi \\ pooling};
\node at (-4.75,-0.25) [fit label] (bv) {Base visual network};
\node at (-2,1) [block, train] (vfc) {\rotatebox{90}{fc layers}};
\node at ( 0,2) [streamblock, test] (vcls){$\bm W_{cls}^v$};
\node at ( 0,0) [streamblock,loc] (vloc) {$\bm W_{loc}^v$};
\node at ( 1.25,0) [transition] (vsig) {$\sigma$};
\node at ( 2,1) [place] (vdot) {$\bullet$};
\node at ( 3.5,1) [transition] (vsum) {$\sum$};
\node at ( 4.5,1) [transition] (l2v) {$\ell_2$ };
\begin{scope}[on background layer]
\node[draw,fit node,inner sep=2mm,bbox, fit=(bv)(conv)(vfc)(vcls)(vloc)(vdot)(vsum)(vsig)(l2v)](vid) {};
\end{scope}
\node[anchor=south east, fit label] at (vid.south east) {Visual Network};

%Audio Net
\node at (-9,-3) [inner sep=0pt] (au) 
{\includegraphics[width=.15\textwidth]{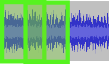}};
\node at (-4.75,-3) [blocknet,fixed] (vggish) {vggish};
\node at (-4.75,-4.25) [fit label] (ba) {Base audio network};
\node at (-2,-3) [block, train] (afc) {\rotatebox{90}{fc layer}};
\node at ( 0,-2) [streamblock, test] (acls){$\bm W_{cls}^a$};
\node at ( 0,-4) [streamblock, loc] (aloc) {$\bm W_{loc}^a$};
\node at ( 1.25,-4) [transition] (asig) {$\sigma$};
\node at ( 2,-3) [place] (adot) {$\bullet$};
\node at ( 3.5,-3) [transition] (asum) {$\sum$};
\node at ( 4.5,-3) [transition] (l2a) {$\ell_2$};

\begin{scope}[on background layer]
\node[draw,fit node,inner sep=2mm,bbox, fit=(ba)(vggish)(afc)(acls)(aloc)(adot)(asum)(asig)(l2a)](aud) {};
\end{scope}
\node[anchor=south east, fit label] at (aud.south east) {Audio network};

\node at (6.0,-1) [place] (gsum) {$+$};

\node at (7.0,-1) [transitioncls] (class) {\rotatebox{90}{\sc classification}};
%All arrows
\draw [->] (vis) --node[above left]{$\mathcal{R}$} (conv);
\draw [->] (vis.north east) -| (roi.east);
\draw [->] (conv) -- (roi);
\draw [->] (roi) --node[above]{$\bm{x}_v$} (vfc);
\draw [->] (vfc.east) -| (-1,2) -- (vcls.west);
\draw [->] (vfc.east) -| (-1,0) -- (vloc.west);
\draw [->] (vcls) -| (vdot);
\draw [->] (vloc) -- (vsig);
\draw [->] (vsig) -| (vdot);
\draw [->] (vdot) -- (vsum);
\draw [->] (vsum) -- (l2v);
\draw [->] (l2v) -| (gsum);

\draw [->] (au) -- node[above left]{$\mathcal{S}$} (vggish);
\draw [->] (vggish) -- node[above]{$\bm x_a$} (afc);
\draw [->] (afc.east) -| (-1,-2) -- (acls.west);
\draw [->] (afc.east) -| (-1,-4) -- (aloc.west);
\draw [->] (acls) -| (adot);
\draw [->] (aloc) -- (asig);
\draw [->] (asig) -| (adot);
\draw [->] (adot) -- (asum);
\draw [->] (asum) -- (l2a);
\draw [->] (l2a) -| (gsum);

\draw [->] (gsum) -- (class);
\end{tikzpicture}}

%% file: eccv2018submission.bbl
\begin{thebibliography}{10}
	
	\bibitem{JiangCottonChangEtAl2009}
	Jiang, W., Cotton, C., Chang, S.F., Ellis, D., Loui, A.:
	\newblock {Short-term audiovisual atoms for generic video concept
		classification}.
	\newblock In: Proceedings of the 17th ACM International Conference on
	Multimedia, ACM (2009)  5--14
	
	\bibitem{ChangEllisJiangEtAl2007}
	Chang, S.F., Ellis, D., Jiang, W., Lee, K., Yanagawa, A., Loui, A.C., Luo, J.:
	\newblock Large-scale multimodal semantic concept detection for consumer video.
	\newblock In: Proceedings of the international workshop on Workshop on
	multimedia information retrieval, ACM (2007)  255--264
	
	\bibitem{jiang2018exploiting}
	Jiang, Y.G., Wu, Z., Wang, J., Xue, X., Chang, S.F.:
	\newblock Exploiting feature and class relationships in video categorization
	with regularized deep neural networks.
	\newblock IEEE transactions on pattern analysis and machine intelligence
	\textbf{40}(2) (2018)  352--364
	
	\bibitem{jiang2013high}
	Jiang, Y.G., Bhattacharya, S., Chang, S.F., Shah, M.:
	\newblock High-level event recognition in unconstrained videos.
	\newblock International journal of multimedia information retrieval
	\textbf{2}(2) (2013)  73--101
	
	\bibitem{IzadiniaSaleemiShah2013}
	Izadinia, H., Saleemi, I., Shah, M.:
	\newblock Multimodal analysis for identification and segmentation of
	moving-sounding objects.
	\newblock Multimedia, IEEE Transactions on \textbf{15}(2) (Feb 2013)  378--390
	
	\bibitem{KidronSchechnerElad2005}
	Kidron, E., Schechner, Y., Elad, M.:
	\newblock Pixels that sound.
	\newblock In: Computer Vision and Pattern Recognition, 2005. CVPR 2005. IEEE
	Computer Society Conference on. Volume~1. (June 2005)  88--95 vol. 1
	
	\bibitem{owens2016visually}
	Owens, A., Isola, P., McDermott, J., Torralba, A., Adelson, E.H., Freeman,
	W.T.:
	\newblock Visually indicated sounds.
	\newblock In: Proceedings of the IEEE Conference on Computer Vision and Pattern
	Recognition. (2016)  2405--2413
	
	\bibitem{owens2016ambient}
	Owens, A., Wu, J., McDermott, J.H., Freeman, W.T., Torralba, A.:
	\newblock Ambient sound provides supervision for visual learning.
	\newblock In: European Conference on Computer Vision, Springer (2016)  801--816
	
	\bibitem{Arandjelovic17}
	Arandjelovi\'c, R., Zisserman, A.:
	\newblock Look, listen and learn.
	\newblock In: IEEE International Conference on Computer Vision. (2017)
	
	\bibitem{Arandjelovic17b}
	Arandjelovi\'c, R., Zisserman, A.:
	\newblock Objects that sound.
	\newblock CoRR \textbf{abs/1712.06651} (2017)
	
	\bibitem{andrew2013deep}
	Andrew, G., Arora, R., Bilmes, J., Livescu, K.:
	\newblock Deep canonical correlation analysis.
	\newblock In: International Conference on Machine Learning. (2013)  1247--1255
	
	\bibitem{dietterich1997solving}
	Dietterich, T.G., Lathrop, R.H., Lozano-P{\'e}rez, T.:
	\newblock Solving the multiple instance problem with axis-parallel rectangles.
	\newblock Artificial intelligence \textbf{89}(1-2) (1997)  31--71
	
	\bibitem{bilen2014weakly}
	Bilen, H., Pedersoli, M., Tuytelaars, T.:
	\newblock Weakly supervised object detection with posterior regularization.
	\newblock In: Proceedings BMVC 2014. (2014)  1--12
	
	\bibitem{kantorov2016contextlocnet}
	Kantorov, V., Oquab, M., Cho, M., Laptev, I.:
	\newblock Contextlocnet: Context-aware deep network models for weakly
	supervised localization.
	\newblock In: European Conference on Computer Vision, Springer (2016)  350--365
	
	\bibitem{bilen2016weakly}
	Bilen, H., Vedaldi, A.:
	\newblock Weakly supervised deep detection networks.
	\newblock In: Proceedings of the IEEE Conference on Computer Vision and Pattern
	Recognition. (2016)  2846--2854
	
	\bibitem{zhang2006multiple}
	Zhang, C., Platt, J.C., Viola, P.A.:
	\newblock Multiple instance boosting for object detection.
	\newblock In: Advances in neural information processing systems. (2006)
	1417--1424
	
	\bibitem{cinbis2017weakly}
	Cinbis, R.G., Verbeek, J., Schmid, C.:
	\newblock Weakly supervised object localization with multi-fold multiple
	instance learning.
	\newblock IEEE transactions on pattern analysis and machine intelligence
	\textbf{39}(1) (2017)  189--203
	
	\bibitem{oquab2015object}
	Oquab, M., Bottou, L., Laptev, I., Sivic, J.:
	\newblock Is object localization for free?-weakly-supervised learning with
	convolutional neural networks.
	\newblock In: Proceedings of the IEEE Conference on Computer Vision and Pattern
	Recognition. (2015)  685--694
	
	\bibitem{zhou2016learning}
	Zhou, B., Khosla, A., Lapedriza, A., Oliva, A., Torralba, A.:
	\newblock Learning deep features for discriminative localization.
	\newblock In: Computer Vision and Pattern Recognition (CVPR), 2016 IEEE
	Conference on, IEEE (2016)  2921--2929
	
	\bibitem{bilen2014object}
	Bilen, H., Namboodiri, V.P., Van~Gool, L.J.:
	\newblock Object and action classification with latent window parameters.
	\newblock International Journal of Computer Vision \textbf{106}(3) (2014)
	237--251
	
	\bibitem{deselaers2010localizing}
	Deselaers, T., Alexe, B., Ferrari, V.:
	\newblock Localizing objects while learning their appearance.
	\newblock In: European conference on computer vision, Springer (2010)  452--466
	
	\bibitem{song2014weakly}
	Song, H.O., Lee, Y.J., Jegelka, S., Darrell, T.:
	\newblock Weakly-supervised discovery of visual pattern configurations.
	\newblock In: Advances in Neural Information Processing Systems. (2014)
	1637--1645
	
	\bibitem{kumar2010self}
	Kumar, M.P., Packer, B., Koller, D.:
	\newblock Self-paced learning for latent variable models.
	\newblock In: Advances in Neural Information Processing Systems. (2010)
	1189--1197
	
	\bibitem{kolesnikov2016seed}
	Kolesnikov, A., Lampert, C.H.:
	\newblock Seed, expand and constrain: Three principles for weakly-supervised
	image segmentation.
	\newblock In: European Conference on Computer Vision, Springer (2016)  695--711
	
	\bibitem{gkioxari2015contextual}
	Gkioxari, G., Girshick, R., Malik, J.:
	\newblock Contextual action recognition with r* cnn.
	\newblock In: Proceedings of the IEEE international conference on computer
	vision. (2015)  1080--1088
	
	\bibitem{zitnick2014edge}
	Zitnick, C.L., Doll{\'a}r, P.:
	\newblock Edge boxes: Locating object proposals from edges.
	\newblock In: European Conference on Computer Vision, Springer (2014)  391--405
	
	\bibitem{uijlings2013selective}
	Uijlings, J.R., Van De~Sande, K.E., Gevers, T., Smeulders, A.W.:
	\newblock Selective search for object recognition.
	\newblock International journal of computer vision \textbf{104}(2) (2013)
	154--171
	
	\bibitem{girshick2015fast}
	Girshick, R.:
	\newblock Fast r-cnn.
	\newblock In: Computer Vision (ICCV), 2015 IEEE International Conference on,
	IEEE (2015)  1440--1448
	
	\bibitem{he2015spatial}
	He, K., Zhang, X., Ren, S., Sun, J.:
	\newblock Spatial pyramid pooling in deep convolutional networks for visual
	recognition.
	\newblock IEEE transactions on pattern analysis and machine intelligence
	\textbf{37}(9) (2015)  1904--1916
	
	\bibitem{mesaros2015sound}
	Mesaros, A., Heittola, T., Dikmen, O., Virtanen, T.:
	\newblock Sound event detection in real life recordings using coupled matrix
	factorization of spectral representations and class activity annotations.
	\newblock In: Acoustics, Speech and Signal Processing (ICASSP), 2015 IEEE
	International Conference on, IEEE (2015)  151--155
	
	\bibitem{zhuang2010real}
	Zhuang, X., Zhou, X., Hasegawa-Johnson, M.A., Huang, T.S.:
	\newblock Real-world acoustic event detection.
	\newblock Pattern Recognition Letters \textbf{31}(12) (2010)  1543--1551
	
	\bibitem{adavanne2017sound}
	Adavanne, S., Pertil{\"a}, P., Virtanen, T.:
	\newblock Sound event detection using spatial features and convolutional
	recurrent neural network.
	\newblock In: Acoustics, Speech and Signal Processing (ICASSP), 2017 IEEE
	International Conference on, IEEE (2017)  771--775
	
	\bibitem{bisot2017overlapping}
	Bisot, V., Essid, S., Richard, G.:
	\newblock Overlapping sound event detection with supervised nonnegative matrix
	factorization.
	\newblock In: Acoustics, Speech and Signal Processing (ICASSP), 2017 IEEE
	International Conference on, IEEE (2017)  31--35
	
	\bibitem{kumar2016audio}
	Kumar, A., Raj, B.:
	\newblock Audio event detection using weakly labeled data.
	\newblock In: Proceedings of the 2016 ACM on Multimedia Conference, ACM (2016)
	1038--1047
	
	\bibitem{gemmeke2017audio}
	Gemmeke, J.F., Ellis, D.P., Freedman, D., Jansen, A., Lawrence, W., Moore,
	R.C., Plakal, M., Ritter, M.:
	\newblock Audio set: An ontology and human-labeled dataset for audio events.
	\newblock In: Acoustics, Speech and Signal Processing (ICASSP), 2017 IEEE
	International Conference on, IEEE (2017)  776--780
	
	\bibitem{DCASE2017challenge}
	A.~Mesaros, T.~Heittola, A.~Diment, B.~Elizalde, A.~Shah, E.~Vincent, B.~Raj,
	and T.~Virtanen.
	\newblock {DCASE} 2017 challenge setup: Tasks, datasets and baseline system.
	\newblock In {\em Proceedings of the Detection and Classification of Acoustic
		Scenes and Events 2017 Workshop (DCASE2017)}, pages 85--92, November 2017.
	
	\bibitem{Xu2017}
	Xu, Y., Kong, Q., Wang, W., Plumbley, M.D.:
	\newblock Surrey-{CVSSP} system for {DCASE2017} challenge task4.
	\newblock Technical report, DCASE2017 Challenge (September 2017)
	
	\bibitem{Salamon2017}
	Salamon, J., McFee, B., Li, P.:
	\newblock {DCASE} 2017 submission: Multiple instance learning for sound event
	detection.
	\newblock Technical report, DCASE2017 Challenge (September 2017)
	
	\bibitem{kumar2017knowledge}
	Kumar, A., Khadkevich, M., Fugen, C.:
	\newblock Knowledge transfer from weakly labeled audio using convolutional
	neural network for sound events and scenes.
	\newblock arXiv preprint arXiv:1711.01369 (2017)
	
	\bibitem{yuhas1989integration}
	Yuhas, B.P., Goldstein, M.H., Sejnowski, T.J.:
	\newblock Integration of acoustic and visual speech signals using neural
	networks.
	\newblock IEEE Communications Magazine \textbf{27}(11) (1989)  65--71
	
	\bibitem{becker1992self}
	Becker, S., Hinton, G.E.:
	\newblock Self-organizing neural network that discovers surfaces in random-dot
	stereograms.
	\newblock Nature \textbf{355}(6356) (1992)  161
	
	\bibitem{aytar2016soundnet}
	Aytar, Y., Vondrick, C., Torralba, A.:
	\newblock Soundnet: Learning sound representations from unlabeled video.
	\newblock In: Advances in Neural Information Processing Systems. (2016)
	892--900
	
	\bibitem{aytar2017see}
	Aytar, Y., Vondrick, C., Torralba, A.:
	\newblock See, hear, and read: Deep aligned representations.
	\newblock arXiv preprint arXiv:1706.00932 (2017)
	
	\bibitem{ngiam2011multimodal}
	Ngiam, J., Khosla, A., Kim, M., Nam, J., Lee, H., Ng, A.Y.:
	\newblock Multimodal deep learning.
	\newblock In: Proceedings of the 28th international conference on machine
	learning (ICML-11). (2011)  689--696
	
	\bibitem{girshick2014rich}
	Girshick, R., Donahue, J., Darrell, T., Malik, J.:
	\newblock Rich feature hierarchies for accurate object detection and semantic
	segmentation.
	\newblock In: Proceedings of the IEEE conference on computer vision and pattern
	recognition. (2014)  580--587
	
	\bibitem{wang2013regionlets}
	Wang, X., Yang, M., Zhu, S., Lin, Y.:
	\newblock Regionlets for generic object detection.
	\newblock In: Computer Vision (ICCV), 2013 IEEE International Conference on,
	IEEE (2013)  17--24
	
	\bibitem{hosang2014good}
	Hosang, J., Benenson, R., Schiele, B.:
	\newblock How good are detection proposals, really?
	\newblock In: 25th British Machine Vision Conference, BMVA Press (2014)  1--12
	
	\bibitem{krizhevsky2012imagenet}
	Krizhevsky, A., Sutskever, I., Hinton, G.E.:
	\newblock Imagenet classification with deep convolutional neural networks.
	\newblock In: Advances in neural information processing systems. (2012)
	1097--1105
	
	\bibitem{deng2009imagenet}
	Deng, J., Dong, W., Socher, R., Li, L.J., Li, K., Fei-Fei, L.:
	\newblock Imagenet: A large-scale hierarchical image database.
	\newblock In: Computer Vision and Pattern Recognition, 2009. CVPR 2009. IEEE
	Conference on, IEEE (2009)  248--255
	
	\bibitem{hershey2017cnn}
	Hershey, S., Chaudhuri, S., Ellis, D.P., Gemmeke, J.F., Jansen, A., Moore,
	R.C., Plakal, M., Platt, D., Saurous, R.A., Seybold, B.,  et~al.:
	\newblock Cnn architectures for large-scale audio classification.
	\newblock In: Acoustics, Speech and Signal Processing (ICASSP), 2017 IEEE
	International Conference on, IEEE (2017)  131--135
	
	\bibitem{45619}
	Abu-El-Haija, S., Kothari, N., Lee, J., Natsev, A.P., Toderici, G.,
	Varadarajan, B., Vijayanarasimhan, S.:
	\newblock Youtube-8m: A large-scale video classification benchmark.
	\newblock In: arXiv:1609.08675. (2016)
	
	\bibitem{adam}
	D.~P. Kingma and J.~Ba.
	\newblock Adam: A method for stochastic optimization.
	\newblock In {ICLR}, 2015.
\end{thebibliography}
